\documentclass{article}

\newif\ifsubmission\submissionfalse
\PassOptionsToPackage{numbers, compress}{natbib}
\ifsubmission
 \usepackage{neurips_2019}
\else
 \usepackage[final]{neurips_2019}
\fi

\usepackage[utf8]{inputenc} 
\usepackage[T1]{fontenc}    
\usepackage{hyperref}       
\usepackage{url}            
\usepackage{booktabs}       
\usepackage{amsfonts}       
\usepackage{nicefrac}       
\usepackage{microtype}      
\usepackage[textsize=tiny,textwidth=8em,backgroundcolor=yellow]{todonotes} 
\usepackage{algorithm}
\usepackage{algorithmic}
\usepackage{setspace}
\usepackage{caption}

\definecolor{mygray}{gray}{0.5}



\usepackage{amsmath,amsfonts,bm}

\def\eqref#1{equation~\ref{#1}}

\def\1{\bm{1}}

\DeclareMathAlphabet{\mathsfit}{\encodingdefault}{\sfdefault}{m}{sl}
\SetMathAlphabet{\mathsfit}{bold}{\encodingdefault}{\sfdefault}{bx}{n}

\newcommand{\pmodel}{{\rm p}_{\rm{model}}}

\DeclareMathOperator{\augment}{Augment}
\DeclareMathOperator{\betadist}{Beta}

\DeclareMathOperator{\sharpen}{Sharpen}
\DeclareMathOperator{\shuffle}{Shuffle}
\DeclareMathOperator{\concat}{Concat}

\DeclareMathOperator{\mixup}{MixUp}
\DeclareMathOperator{\mixmatch}{MixMatch}
\DeclareMathOperator{\xent}{H}

\usepackage{cleveref}

\title{MixMatch: A Holistic Approach to \\Semi-Supervised Learning}

\author{%
  David Berthelot \\
  Google Research \\
  \texttt{dberth@google.com} \\
  \And
  Nicholas Carlini \\
  Google Research \\
  \texttt{ncarlini@google.com} \\
  \And
  Ian Goodfellow \\
  Work done at Google \\
  \texttt{ian-academic@mailfence.com} \\
  \And
  Avital Oliver \\
  Google Research \\
  \texttt{avitalo@google.com} \\
  \And
  Nicolas Papernot \\
  Google Research \\
  \texttt{papernot@google.com} \\
  \And
  Colin Raffel \\
  Google Research \\
  \texttt{craffel@google.com}
}

\begin{document}

\maketitle

\begin{abstract}
Semi-supervised learning has proven to be a powerful paradigm for leveraging unlabeled data to mitigate the reliance on large labeled datasets.
In this work, we unify the current dominant approaches for semi-supervised learning to produce a new algorithm, $\mixmatch$, that
guesses low-entropy labels for data-augmented unlabeled examples and mixes labeled and unlabeled data using $\mixup$.
$\mixmatch$ obtains state-of-the-art results by a large margin across many datasets and labeled data amounts. For example,
on CIFAR-10 with 250 labels, we reduce error rate by a factor of 4 (from $38\%$ to $11\%$) and by a factor of 2 on STL-10.
We also demonstrate how $\mixmatch$ can help achieve a dramatically better accuracy-privacy trade-off for differential privacy.
Finally, we perform an ablation study to tease apart which components of $\mixmatch$ are most important for its success.
We release all code used in our experiments.\footnote{\url{https://github.com/google-research/mixmatch}}
\end{abstract}

\section{Introduction}

Much of the recent success in training large, deep neural networks is thanks in part to the existence of large labeled datasets.
Yet, collecting labeled data is expensive for many learning tasks because it necessarily involves expert knowledge.
This is perhaps best illustrated by medical tasks where measurements call for expensive machinery and labels are the fruit of a time-consuming analysis that draws from multiple human experts.
Furthermore, data labels may contain private information.
In comparison, in many tasks it is much easier or cheaper to obtain unlabeled data.

Semi-supervised learning \cite{chapelle2006semi} (SSL) seeks to largely alleviate the need for labeled data by allowing a model to leverage unlabeled data.
Many recent approaches for semi-supervised learning add a loss term which is computed on unlabeled data and encourages the model to generalize better to unseen data.
In much recent work, this loss term falls into one of three classes (discussed further in \Cref{sec:related_work}): entropy minimization \cite{grandvalet2005semi,lee2013pseudo}---which encourages the model to output confident predictions on unlabeled data; consistency regularization---which encourages the model to produce the same output distribution when its inputs are perturbed; and generic regularization---which encourages the model to generalize well and avoid overfitting the training data.

In this paper, we introduce $\mixmatch$, an SSL algorithm which introduces a single loss that gracefully unifies these dominant approaches to semi-supervised learning.
Unlike previous methods, $\mixmatch$ targets all the properties at once which we find leads to the following benefits:

\begin{itemize}
    \item Experimentally, we show that $\mixmatch$ obtains state-of-the-art results on all standard image benchmarks (\cref{sec:ssl_experiments}), and reducing the error rate on CIFAR-10 by a factor of 4;
    \item We further show in an ablation study that $\mixmatch$ is greater than the sum of its parts;
    \item We demonstrate in \cref{sec:dp_experiments} that $\mixmatch$ is useful for differentially private learning, enabling students in the PATE framework~\cite{papernot2016semi} to obtain new state-of-the-art results that simultaneously strengthen both privacy guarantees and accuracy.
\end{itemize}

In short, $\mixmatch$ introduces a unified loss term for unlabeled data that seamlessly reduces entropy while maintaining consistency and remaining compatible with traditional regularization techniques.

\begin{figure}
    \centering
    \includegraphics[width=0.8\textwidth]{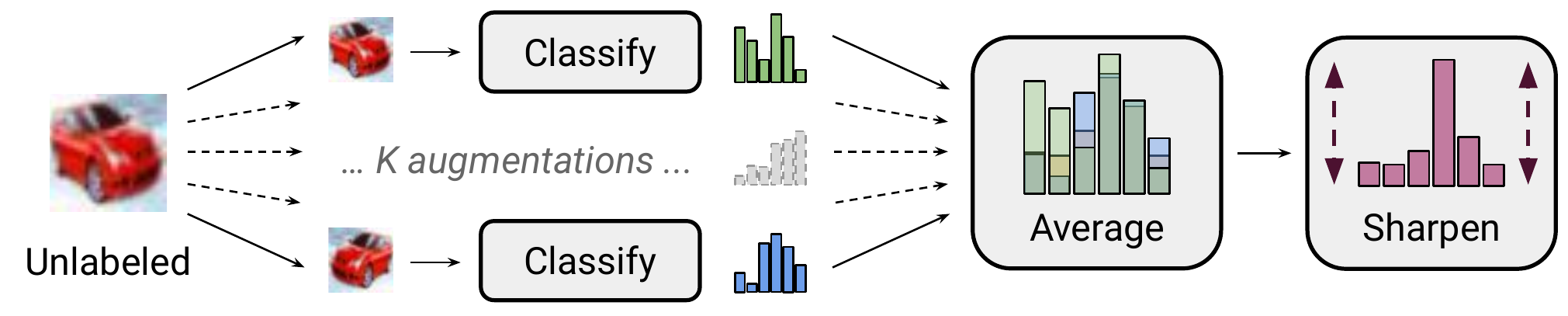}
    \caption{Diagram of the label guessing process used in $\mixmatch$. Stochastic data augmentation is applied to an unlabeled image $K$ times, and each augmented image is fed through the classifier. Then, the average of these $K$ predictions is ``sharpened'' by adjusting the distribution's temperature. See \cref{alg:mixmatch} for a full description.}
    \label{fig:guess_label}
\end{figure}

\section{Related Work}
\label{sec:related_work}

To set the stage for $\mixmatch$, we first introduce existing methods for SSL.
We focus mainly on those which are currently state-of-the-art and that $\mixmatch$ builds on; there is a wide literature on SSL techniques that we do not discuss here (e.g.,\ ``transductive'' models \cite{gammerman1998learning,joachims2003transductive,joachims1999transductive}, graph-based methods \cite{zhu2003semi,bengio2006label,liu2018deep}, generative modeling \cite{Belkin+Niyogi-2002,LasserreJ2006,Russ+Geoff-nips-2007,Coates2011b,Goodfellow2011,kingma2014semi,pu2016variational,odena2016semi,salimans2016improved}, etc.).
More comprehensive overviews are provided in \cite{zhu2003semi,chapelle2006semi}.
In the following, we will refer to a generic model $\pmodel(y \mid x; \theta)$ which produces a distribution over class labels $y$ for an input $x$ with parameters $\theta$.

\subsection{Consistency Regularization}
\label{sec:consistency}

A common regularization technique in supervised learning is \textit{data augmentation}, which applies input transformations assumed to leave class semantics unaffected.
For example, in image classification, it is common to elastically deform or add noise to an input image, which can dramatically change the pixel content of an image without altering its label \cite{ciresan2010deep,simard2003best,cubuk2018autoaugment}.
Roughly speaking, this can artificially expand the size of a training set by generating a near-infinite stream of new, modified data.
\textit{Consistency regularization} applies data augmentation to semi-supervised learning by leveraging the idea that a classifier should output the same class distribution for an unlabeled example even after it has been augmented.
More formally, consistency regularization enforces that an unlabeled example $x$ should be classified the same as $\augment(x)$, an augmentation of itself.

In the simplest case, for unlabeled points $x$, prior work \cite{laine2016temporal,sajjadi2016regularization} adds the loss term
\begin{equation}
\|\pmodel(y \mid \augment(x); \theta) - \pmodel(y \mid \augment(x); \theta)\|_2^2.
  \label{eq:pi_model}
\end{equation}
Note that $\augment(x)$ is a stochastic transformation, so the two terms in \cref{eq:pi_model} are not identical.
``Mean Teacher'' \cite{tarvainen2017weight} replaces one of the terms in \cref{eq:pi_model} with the output of the model using an exponential moving average of model parameter values.
This provides a more stable target and was found empirically to significantly improve results.
A drawback to these approaches is that they use domain-specific data augmentation strategies.
``Virtual Adversarial Training'' \cite{miyato2018virtual} (VAT) addresses this by instead computing an additive perturbation to apply to the input which maximally changes the output class distribution.
MixMatch utilizes a form of consistency regularization through the use of standard data augmentation for images (random horizontal flips and crops).

\subsection{Entropy Minimization}
\label{sec:entmin}

A common underlying assumption in many semi-supervised learning methods is that the classifier's decision boundary should not pass through high-density regions of the marginal data distribution.
One way to enforce this is to require that the classifier output low-entropy predictions on unlabeled data.
This is done explicitly in \cite{grandvalet2005semi} with a loss term which minimizes the entropy of $\pmodel(y \mid x; \theta)$ for unlabeled data $x$.
This form of entropy minimization was combined with VAT in \cite{miyato2018virtual} to obtain stronger results.
``Pseudo-Label'' \cite{lee2013pseudo} does entropy minimization implicitly by constructing hard (1-hot) labels from high-confidence predictions on unlabeled data and using these as training targets in a standard cross-entropy loss.
MixMatch also implicitly achieves entropy minimization through the use of a ``sharpening'' function on the target distribution for unlabeled data, described in \cref{sec:label_guessing}.

\subsection{Traditional Regularization}
Regularization refers to the general approach of imposing a constraint on a model to make it harder to memorize the training data and therefore hopefully make it generalize better to unseen data \cite{hinton1993keeping}.
We use weight decay which penalizes the $L_2$ norm of the model parameters \cite{loshchilov2017fixing,zhang2018three}.
We also use $\mixup$ \cite{zhang2017mixup} in $\mixmatch$ to encourage convex behavior ``between'' examples.
We utilize $\mixup$ as both as a regularizer (applied to labeled datapoints) and a semi-supervised learning method (applied to unlabeled datapoints).
$\mixup$ has been previously applied to semi-supervised learning; in particular, the concurrent work of \cite{verma2019interpolation} uses a subset of the methodology used in MixMatch.
We clarify the differences in our ablation study (\cref{sec:ablation}).

\section{MixMatch}
\label{sec:mixmatch}

In this section, we introduce $\mixmatch$, our proposed semi-supervised learning method.
$\mixmatch$ is a ``holistic'' approach which incorporates ideas and components from the dominant paradigms for SSL discussed in \cref{sec:related_work}.
Given a batch $\mathcal{X}$ of labeled examples with one-hot targets (representing one of $L$ possible labels) and an equally-sized batch $\mathcal{U}$ of unlabeled examples, $\mixmatch$ produces a processed batch of augmented labeled examples $\mathcal{X}^\prime$ and a batch of augmented unlabeled examples with ``guessed'' labels $\mathcal{U}^\prime$.
$\mathcal{U}^\prime$ and $\mathcal{X}^\prime$ are then used in computing separate labeled and unlabeled loss terms.
More formally, the combined loss $\mathcal{L}$ for semi-supervised learning is defined as
\begin{align}
    \mathcal{X}^\prime, \mathcal{U}^\prime &= \mixmatch(\mathcal{X}, \mathcal{U}, T, K, \alpha)\\
    \mathcal{L}_\mathcal{X} &= \frac{1}{|\mathcal{X}^\prime|} \sum_{x, p \in \mathcal{X}^\prime} \xent(p, \pmodel(y \mid x; \theta)) \label{eqn:l_x} \\
    \mathcal{L}_\mathcal{U} &= \frac{1}{L|\mathcal{U}^\prime|} \sum_{u, q \in \mathcal{U}^\prime} \|q - \pmodel(y \mid u; \theta)\|_2^2 \label{eqn:l_u} \\
    \mathcal{L} &= \mathcal{L}_\mathcal{X} + \lambda_\mathcal{U} \mathcal{L}_\mathcal{U} \label{eqn:l_combined}
\end{align}
where $\xent(p, q)$ is the cross-entropy between distributions $p$ and $q$, and $T$, $K$, $\alpha$, and $\lambda_\mathcal{U}$ are hyperparameters described below.
The full $\mixmatch$ algorithm is provided in \cref{alg:mixmatch}, and a diagram of the label guessing process is shown in \cref{fig:guess_label}.
Next, we describe each part of $\mixmatch$.

\begin{algorithm*}[t]
        \caption{$\mixmatch$ takes a batch of labeled data $\mathcal{X}$ and a batch of unlabeled data $\mathcal{U}$ and produces a collection $\mathcal{X}^\prime$ (resp. $\mathcal{U}^\prime$) of processed labeled examples (resp. unlabeled with guessed labels).}
   \label{alg:mixmatch}
\begin{algorithmic}[1]
   \footnotesize
   \STATE {\bfseries Input:} Batch of labeled examples and their one-hot labels $\mathcal{X} = \big((x_b, p_b); b \in (1, \ldots, B)\big)$, batch of unlabeled examples $\mathcal{U} = \big(u_b; b \in (1, \ldots, B)\big)$, sharpening temperature $T$, number of augmentations $K$, $\betadist$ distribution parameter $\alpha$ for $\mixup$.
   \FOR{$b = 1$ \TO $B$}
   \STATE $\hat{x}_b = \augment(x_b)$ \COMMENT{\textit{Apply data augmentation to $x_b$}} \label{line:augment_labeled} \\
   \FOR{$k = 1$ \TO $K$}
   \STATE $\hat{u}_{b, k} = \augment(u_b)$ \COMMENT{\textit{Apply $k^{th}$ round of data augmentation to $u_b$}} \label{line:augment_unlabeled} \\
   \ENDFOR
   \STATE $\bar{q}_b = \frac{1}{K}\sum_k \pmodel(y \mid \hat{u}_{b, k}; \theta)$ \COMMENT{\textit{Compute average predictions across all augmentations of $u_b$}} \label{line:average_prediction} \\
   \STATE $q_b = \sharpen(\bar{q}_b, T)$ \COMMENT{\textit{Apply temperature sharpening to the average prediction (see \cref{eqn:sharpen})}} \label{line:sharpen} \\
   \ENDFOR
   \STATE $\hat{\mathcal{X}} = \big((\hat{x}_b, p_b); b \in (1, \ldots, B)\big)$ \COMMENT{\textit{Augmented labeled examples and their labels}} \label{line:hat_x} \\
   \STATE $\hat{\mathcal{U}} = \big((\hat{u}_{b, k}, q_b); b \in (1, \ldots, B), k \in (1, \ldots, K)\big)$ \COMMENT{\textit{Augmented unlabeled examples, guessed labels}} \label{line:hat_u} \\
   \STATE $\mathcal{W} = \shuffle\mathopen{}\big(\mathopen{}\concat(\hat{\mathcal{X}}, \hat{\mathcal{U}})\big)$ \COMMENT{\textit{Combine and shuffle labeled and unlabeled data}} \label{line:w} \\
   \STATE $\mathcal{X}^\prime = \big(\mathopen{}\mixup(\hat{\mathcal{X}}_i, \mathcal{W}_i); i \in (1, \ldots, |\hat{\mathcal{X}}|)\big)$ \COMMENT{Apply \textit{$\mixup$ to labeled data and entries from $\mathcal{W}$}} \label{line:x_prime} \\
   \STATE $\mathcal{U}^\prime = \big(\mathopen{}\mixup(\hat{\mathcal{U}}_i, \mathcal{W}_{i+|\hat{\mathcal{X}}|}); i \in (1, \ldots, |\hat{\mathcal{U}}|)\big)$ \COMMENT{\textit{Apply $\mixup$ to unlabeled data and the rest of $\mathcal{W}$}} \label{line:u_prime} \\
   \RETURN $\mathcal{X}^\prime, \mathcal{U}^\prime$
\end{algorithmic}
\end{algorithm*}

\subsection{Data Augmentation}

As is typical in many SSL methods, we use data augmentation both on labeled and unlabeled data.
For each $x_b$ in the batch of labeled data $\mathcal{X}$, we generate a transformed version $\hat{x}_b = \augment(x_b)$ (\cref{alg:mixmatch}, line \ref{line:augment_labeled}).
For each $u_b$ in the batch of unlabeled data $\mathcal{U}$, we generate $K$ augmentations $\hat{u}_{b, k} = \augment(u_b), k \in (1, \ldots, K)$ (\cref{alg:mixmatch}, line \ref{line:augment_unlabeled}).
We use these individual augmentations to generate a ``guessed label'' $q_b$ for each $u_b$, through a process we describe in the following subsection.

\subsection{Label Guessing}
\label{sec:label_guessing}

For each unlabeled example in $\mathcal{U}$, $\mixmatch$ produces a ``guess'' for the example's label using the model's predictions.
This guess is later used in the unsupervised loss term.
To do so, we compute the average of the model's predicted class distributions across all the $K$ augmentations of $u_b$ by
\begin{equation}
\bar{q}_b = \frac{1}{K}\sum_{k = 1}^K \pmodel(y \mid \hat{u}_{b, k}; \theta)
\end{equation}
in \cref{alg:mixmatch}, line \ref{line:average_prediction}.
Using data augmentation to obtain an artificial target for an unlabeled example is common in consistency regularization methods \cite{laine2016temporal,sajjadi2016regularization,tarvainen2017weight}.

\paragraph{Sharpening.} In generating a label guess, we perform one additional step inspired by the success of entropy minimization in semi-supervised learning (discussed in \cref{sec:entmin}).
Given the average prediction over augmentations $\bar{q}_b$, we apply a sharpening function to reduce the entropy of the label distribution.
In practice, for the sharpening function, we use the common approach of adjusting the ``temperature'' of this categorical distribution \cite{goodfellow2016deep}, which is defined as the operation
\begin{equation}
    \sharpen(p, T)_i := p_i^{\frac{1}{T}}\bigg/ \sum_{j = 1}^L p_j^{\frac{1}{T}}
    \label{eqn:sharpen}
\end{equation}
where $p$ is some input categorical distribution (specifically in $\mixmatch$, $p$ is the average class prediction over augmentations $\bar{q}_b$, as shown in \cref{alg:mixmatch}, line \ref{line:sharpen}) and $T$ is a hyperparameter.
As $T \rightarrow 0$, the output of $\sharpen(p, T)$ will approach a Dirac (``one-hot'') distribution.
Since we will later use $q_b = \sharpen(\bar{q}_b, T)$ as a target for the model's prediction for an augmentation of $u_b$, lowering the temperature encourages the model to produce lower-entropy predictions.

\subsection{MixUp}

We use $\mixup$ for semi-supervised learning, and unlike past work for SSL we mix both labeled examples and unlabeled examples with label guesses (generated as described in \cref{sec:label_guessing}).
To be compatible with our separate loss terms, we define a slightly modified version of $\mixup$.
For a pair of two examples with their corresponding labels probabilities $(x_1, p_1), (x_2, p_2)$ we compute $(x^\prime, p^\prime)$ by
\begin{align}
   \lambda &\sim \betadist(\alpha, \alpha) \\
   \lambda^\prime &= \max(\lambda, 1 - \lambda) \label{eqn:lambda_prime} \\
   x^\prime &= \lambda^\prime x_1 + (1 - \lambda^\prime)x_2 \\
   p^\prime &= \lambda^\prime p_1 + (1 - \lambda^\prime)p_2
\end{align}
where $\alpha$ is a hyperparameter.
Vanilla $\mixup$ omits \cref{eqn:lambda_prime} (i.e.\ it sets $\lambda^\prime = \lambda$).
Given that labeled and unlabeled examples are concatenated in the same batch, we need to preserve the order of the batch to compute individual loss components appropriately.
This is achieved by \cref{eqn:lambda_prime} which ensures that $x^\prime$ is closer to $x_1$ than to $x_2$.
To apply $\mixup$, we first collect all augmented labeled examples with their labels and all unlabeled examples with their guessed labels into
\begin{align}
    \hat{\mathcal{X}} & = \big((\hat{x}_b, p_b); b \in (1, \ldots, B)\big) \\
    \hat{\mathcal{U}} & = \big((\hat{u}_{b, k}, q_b); b \in (1, \ldots, B), k \in (1, \ldots, K)\big)
\end{align} (\cref{alg:mixmatch}, lines \ref{line:hat_x}--\ref{line:hat_u}).
Then, we combine these collections and shuffle the result to form $\mathcal{W}$ which will serve as a data source for $\mixup$ (\cref{alg:mixmatch}, line \ref{line:w}).
For each the $i^{th}$ example-label pair in $\hat{\mathcal{X}}$, we compute $\mixup(\hat{\mathcal{X}}_i, \mathcal{W}_i)$ and add the result to the collection $\mathcal{X}^\prime$ (\cref{alg:mixmatch}, line \ref{line:x_prime}).
We compute $\mathcal{U}^\prime_i = \mixup(\hat{\mathcal{U}}_i, \mathcal{W}_{i+|\hat{\mathcal{X}}|})$ for $i \in (1, \ldots, |\hat{\mathcal{U}}|)$, intentionally using the remainder of $\mathcal{W}$ that was not used in the construction of $\mathcal{X}^\prime$ (\cref{alg:mixmatch}, line \ref{line:u_prime}).
To summarize, $\mixmatch$ transforms $\mathcal{X}$ into $\mathcal{X}^\prime$, a collection of labeled examples which have had data augmentation and $\mixup$ (potentially mixed with an unlabeled example) applied.
Similarly, $\mathcal{U}$ is transformed into $\mathcal{U}^\prime$, a collection of multiple augmentations of each unlabeled example with corresponding label guesses.

\subsection{Loss Function}
\label{sec:loss_function}

Given our processed batches $\mathcal{X}^\prime$ and $\mathcal{U}^\prime$, we use the standard semi-supervised loss shown in \cref{eqn:l_x,eqn:l_u,eqn:l_combined}.
\Cref{eqn:l_combined} combines the typical cross-entropy loss between labels and model predictions from $\mathcal{X}^\prime$ with the squared $L_2$ loss on predictions and guessed labels from $\mathcal{U}^\prime$.
We use this $L_2$ loss in \cref{eqn:l_u} (the multiclass Brier score \cite{brier1950verification})  because, unlike the cross-entropy, it is bounded and less sensitive to incorrect predictions.
For this reason, it is often used as the unlabeled data loss in SSL \cite{laine2016temporal,tarvainen2017weight} as well as a measure of predictive uncertainty \cite{lakshminarayanan2017simple}.
We do not propagate gradients through computing the guessed labels, as is standard \cite{laine2016temporal,tarvainen2017weight,miyato2018virtual,oliver2018realistic}

\subsection{Hyperparameters}

Since $\mixmatch$ combines multiple mechanisms for leveraging unlabeled data, it introduces various hyperparameters -- specifically, the sharpening temperature $T$, number of unlabeled augmentations $K$, $\alpha$ parameter for $\betadist$ in $\mixup$, and the unsupervised loss weight $\lambda_{\mathcal{U}}$.
In practice, semi-supervised learning methods with many hyperparameters can be problematic because cross-validation is difficult with small validation sets \cite{oliver2018realistic,rasmus2015semi,oliver2018realistic}.
However, we find in practice that most of $\mixmatch$'s hyperparameters can be fixed and do not need to be tuned on a per-experiment or per-dataset basis.
Specifically, for all experiments we set $T = 0.5$ and $K = 2$.
Further, we only change $\alpha$ and $\lambda_{\mathcal{U}}$ on a per-dataset basis; we found that $\alpha=0.75$ and $\lambda_{\mathcal{U}}=100$ are good starting points for tuning.
In all experiments, we linearly ramp up $\lambda_{\mathcal{U}}$ to its maximum value over the first $16{,}000$ steps of training as is common practice \cite{tarvainen2017weight}.

\section{Experiments}
We test the effectiveness of $\mixmatch$ on standard SSL benchmarks (\cref{sec:ssl_experiments}). Our  ablation study teases apart the contribution of each of $\mixmatch$'s components (\cref{sec:ablation}).
As an additional application, we consider privacy-preserving learning in \cref{sec:dp_experiments}.

\subsection{Implementation details}
\label{sec:implementation}

Unless otherwise noted, in all experiments we use the ``Wide ResNet-28'' model from \cite{oliver2018realistic}.
Our implementation of the model and training procedure closely matches that of \cite{oliver2018realistic} (including using 5000 examples to select the hyperparameters), except for the following differences:
First, instead of decaying the learning rate, we evaluate models using an exponential moving average of their parameters with a decay rate of $0.999$.
Second, we apply a weight decay of $0.0004$ at each update for the Wide ResNet-28 model.
Finally, we checkpoint every $2^{16}$ training samples and report the median error rate of the last 20 checkpoints.
This simplifies the analysis at a potential cost to accuracy by,
for example, averaging checkpoints \cite{athiwaratkun2018improving} or choosing the checkpoint with the lowest validation error.

\subsection{Semi-Supervised Learning}
\label{sec:ssl_experiments}

First, we evaluate the effectiveness of $\mixmatch$ on four standard benchmark datasets: CIFAR-10 and CIFAR-100 \cite{krizhevsky2009learning}, SVHN \cite{netzer2011reading}, and STL-10 \cite{coates2011analysis}.
Standard practice for evaluating semi-supervised learning on the first three datasets is to treat most of the dataset as unlabeled and use a small portion as labeled data.
STL-10 is a dataset specifically designed for SSL, with 5,000 labeled images and 100,000 unlabeled images which are drawn from a slightly different distribution than the labeled data.

\subsubsection{Baseline Methods}

As baselines, we consider the four methods considered in \cite{oliver2018realistic} ($\Pi$-Model \cite{laine2016temporal,sajjadi2016regularization}, Mean Teacher \cite{tarvainen2017weight}, Virtual Adversarial Training \cite{miyato2018virtual}, and Pseudo-Label \cite{lee2013pseudo}) which are described in \cref{sec:related_work}.
We also use $\mixup$ \cite{zhang2017mixup} on its own as a baseline.
$\mixup$ is designed as a regularizer for supervised learning, so we modify it for SSL by applying it both to augmented labeled examples and augmented unlabeled examples with their corresponding predictions.
In accordance with standard usage of $\mixup$, we use a cross-entropy loss between the $\mixup$-generated guess label and the model's prediction.
As advocated by \cite{oliver2018realistic}, we reimplemented each of these methods in the same codebase and applied them to the same model (described in \cref{sec:implementation}) to ensure a fair comparison.
We re-tuned the hyperparameters for each baseline method, which generally resulted in a marginal accuracy improvement compared to those in \cite{oliver2018realistic}, thereby providing a more competitive experimental setting for testing out $\mixmatch$.

\begin{figure}[t]
  \begin{minipage}[l]{\textwidth}
    \begin{minipage}[t]{0.48\textwidth}
      \centering
      \centerline{\includegraphics[width=\columnwidth]{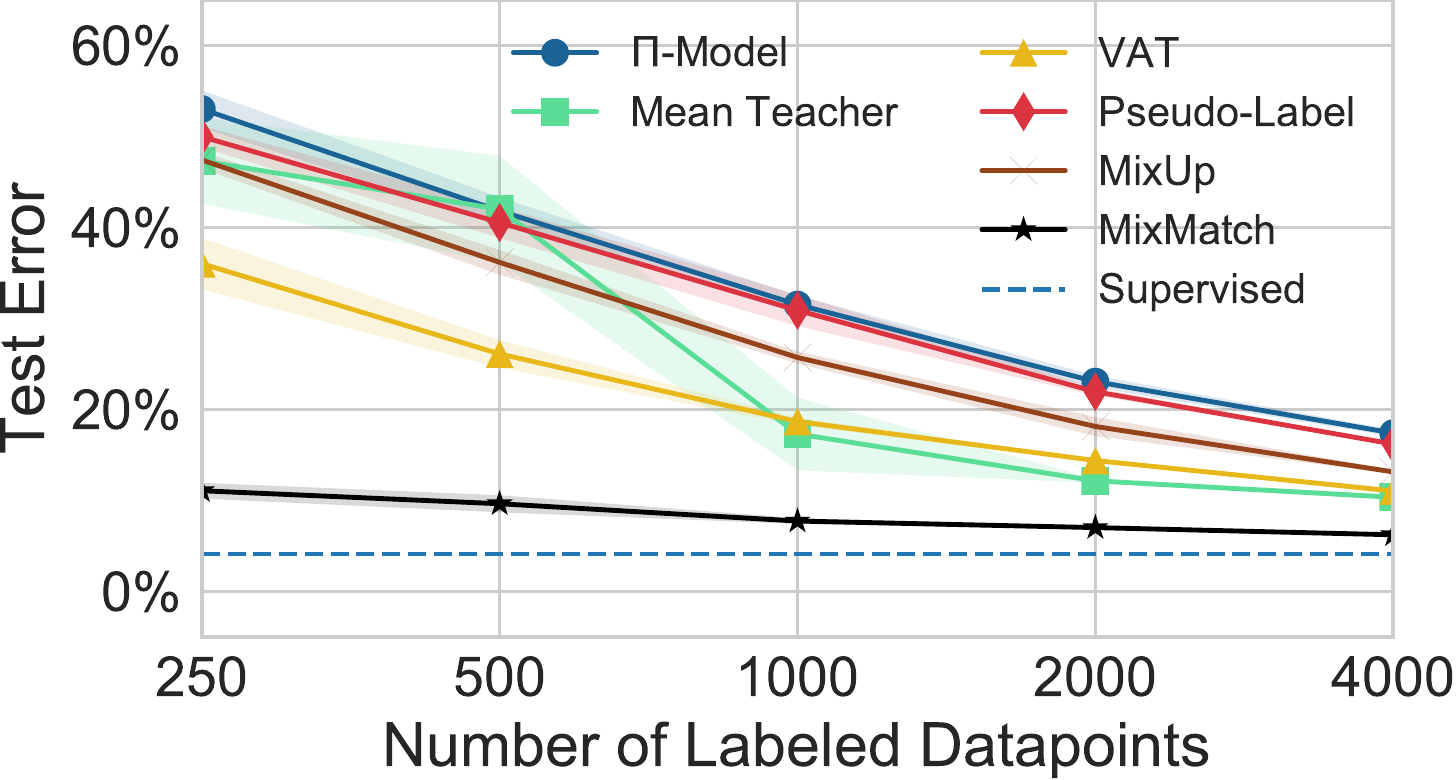}}
      \captionof{figure}{
        Error rate comparison of $\mixmatch$ to baseline methods on CIFAR-10 for a varying number of labels.
        Exact numbers are provided in \cref{tab:cifar10} (appendix).
        ``Supervised'' refers to training with all $50000$ training examples and no unlabeled data.
        With $250$ labels $\mixmatch$ reaches an error rate comparable to next-best method's performance with $4000$ labels.}
      \label{fig:vary_cifar10}
    \end{minipage}\hfill
    \begin{minipage}[t]{0.48\textwidth}
      \centering
      \centerline{\includegraphics[width=\columnwidth]{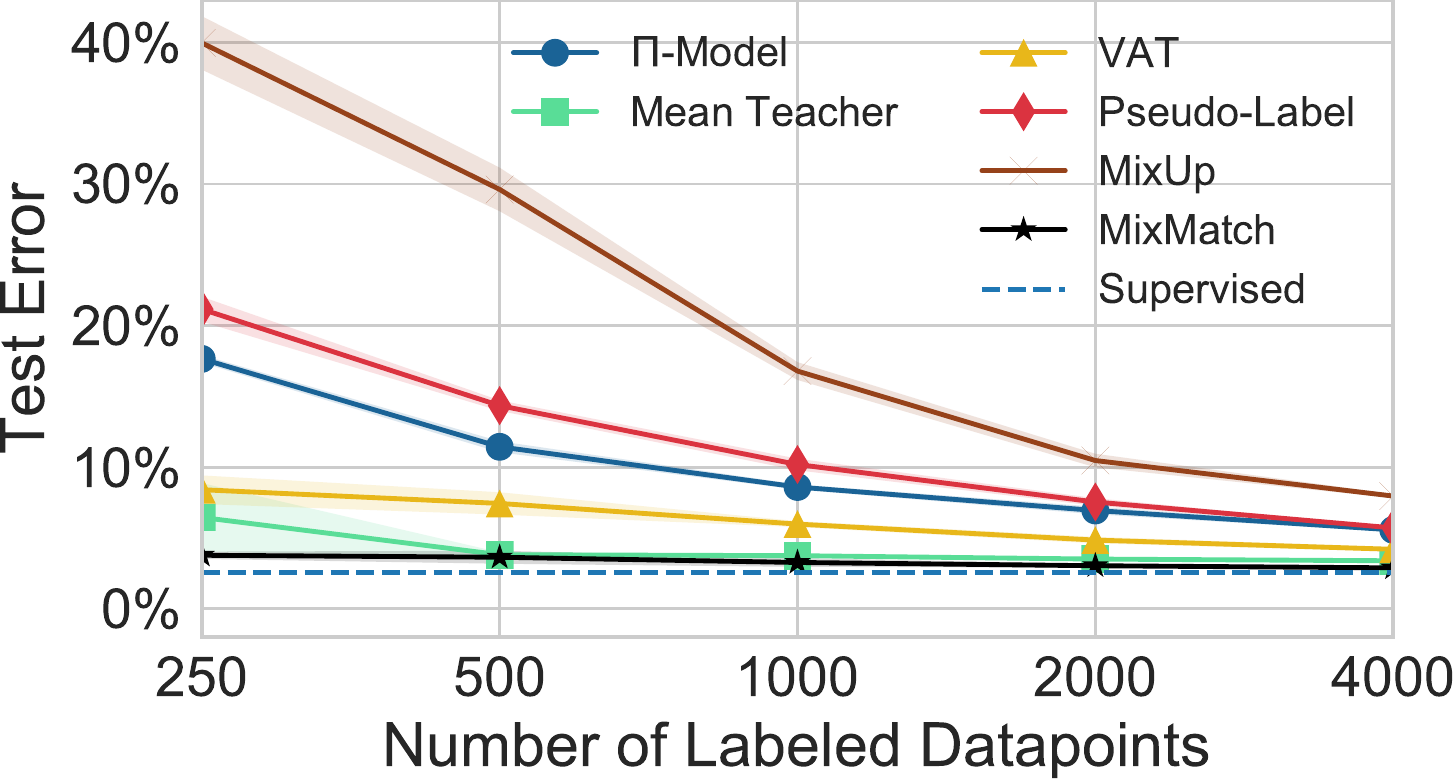}}
      \captionof{figure}{
        Error rate comparison of $\mixmatch$ to baseline methods on SVHN for a varying number of labels.
        Exact numbers are provided in \cref{tab:svhn} (appendix).
        ``Supervised'' refers to training with all $73257$ training examples and no unlabeled data.
        With $250$ examples $\mixmatch$ nearly reaches the accuracy of supervised training for this model.}
    \label{fig:vary_svhn}
\end{minipage}
  \end{minipage}
\end{figure}

\subsubsection{Results}

\paragraph{CIFAR-10} For CIFAR-10, we evaluate the accuracy of each method with a varying number of labeled examples from $250$ to $4000$ (as is standard practice).
The results can be seen in \cref{fig:vary_cifar10}.
We used $\lambda_{\mathcal{U}}=75$ for CIFAR-10.
We created 5 splits for each number of labeled points, each with a different random seed.
Each model was trained on each split and the error rates were reported by the mean and variance across splits.
We find that $\mixmatch$ outperforms all other methods by a significant margin, for example reaching an error rate of $6.24\%$ with $4000$ labels.
For reference, on the same model, fully supervised training on all $50000$ samples achieves an error rate of $4.17\%$.
Furthermore, $\mixmatch$ obtains an error rate of $11.08\%$ with only $250$ labels.
For comparison, at $250$ labels the next-best-performing method (VAT \cite{miyato2018virtual}) achieves an error rate of $36.03$, over $4.5\times$ higher than $\mixmatch$ considering that $4.17\%$ is the error limit obtained on our model with fully supervised learning.
In addition, at $4000$ labels the next-best-performing method (Mean Teacher \cite{tarvainen2017weight}) obtains an error rate of $10.36\%$, which suggests that $\mixmatch$ can achieve similar performance with only $1/16$ as many labels.
We believe that the most interesting comparisons are with very few labeled data points since it reveals the method's sample efficiency which is central to SSL. 

\paragraph{CIFAR-10 and CIFAR-100 with a larger model} Some prior work \cite{tarvainen2017weight,athiwaratkun2018improving} has also considered the use of a larger, $26$ million-parameter model.
Our base model, as used in \cite{oliver2018realistic}, has only $1.5$ million parameters which confounds comparison with these results.
For a more reasonable comparison to these results, we measure the effect of increasing the width of our base ResNet model and evaluate $\mixmatch$'s performance on a 28-layer Wide Resnet model which has $135$ filters per layer, resulting in $26$ million parameters.
We also evaluate $\mixmatch$ on this larger model on CIFAR-100 with $10000$ labels, to compare to the corresponding result from \cite{athiwaratkun2018improving}.
The results are shown in \cref{tab:large_model}.
In general, $\mixmatch$ matches or outperforms the best results from \cite{athiwaratkun2018improving}, though we note that the comparison still remains problematic due to the fact that the model from \cite{tarvainen2017weight,athiwaratkun2018improving} also uses more sophisticated ``shake-shake'' regularization \cite{gastaldi2017shake}.
For this model, we used a weight decay of $0.0008$.
We used $\lambda_{\mathcal{U}} = 75$ for CIFAR-10 and $\lambda_{\mathcal{U}} = 150$ for CIFAR-100.

\begin{table}
  \parbox{.5\linewidth}{
    \centering
    \small
    \begin{tabular}{lrr}
    \toprule
      Method & CIFAR-10 & CIFAR-100 \\
      \midrule
      Mean Teacher \cite{tarvainen2017weight} & $6.28$ & -\\
      SWA \cite{athiwaratkun2018improving}    & $5.00$ & $28.80$  \\
      \midrule
      $\mixmatch$    & $4.95 \pm 0.08$ & $25.88 \pm 0.30$ \\
    \bottomrule
    \end{tabular}
    \vskip 0.1in
    \caption{CIFAR-10 and CIFAR-100 error rate (with $4{,}000$ and $10{,}000$ labels respectively) with larger models ($26$ million parameters).}
    \label{tab:large_model}
  }
  \hfill
  \parbox{.46\linewidth}{
    \centering
    \small
    \begin{tabular}{lrr}
    \toprule
      Method & $1000$ labels & $5000$ labels \\
      \midrule
      CutOut \cite{devries2017improved} & - & $12.74$ \\
      IIC \cite{ji2018invariant} & -  & $11.20$ \\
      SWWAE \cite{zhao2015stacked} & $25.70$ & - \\
      CC-GAN$^2$ \cite{denton2016semi} & $22.20$ & - \\
      \midrule
      $\mixmatch$    & $10.18 \pm 1.46$ & 5.59 \\
    \bottomrule
    \end{tabular}
    \vskip 0.1in
    \caption{STL-10 error rate using $1000$-label splits or the entire $5000$-label training set.}
    \label{tab:stl10}
  }
  \vskip -0.2in
\end{table}

\paragraph{SVHN and SVHN+Extra}
As with CIFAR-10, we evaluate the performance of each SSL method on SVHN with a varying number of labels from $250$ to $4000$.
As is standard practice, we first consider the setting where the $73257$-example training set is split into labeled and unlabeled data.
The results are shown in \cref{fig:vary_svhn}.
We used $\lambda_{\mathcal{U}}=250$.
Here again the models were evaluated on 5 splits for each number of labeled points, each with a different random seed.
We found $\mixmatch$'s performance to be relatively constant (and better than all other methods) across all amounts of labeled data.
Surprisingly, after additional tuning we were able to obtain extremely good performance from Mean Teacher \cite{tarvainen2017weight}, though its error rate was consistently slightly higher than $\mixmatch$'s.

Note that SVHN has two training sets: \textit{train} and \textit{extra}.
In fully-supervised learning, both sets are concatenated to form the full training set ($604388$ samples).
In SSL, for historical reasons the \textit{extra} set was left aside and only train was used ($73257$ samples).
We argue that leveraging both \textit{train} and \textit{extra} for the unlabeled data is more interesting since it exhibits a higher ratio of unlabeled samples over labeled ones.
We report error rates for both SVHN and SVHN+Extra in \cref{tab:svhn_nox}.
For SVHN+Extra we used $\alpha=0.25, \lambda_{\mathcal{U}}=250$ and a lower weight decay of $0.000002$ due to the larger amount of available data.
We found that on both training sets, $\mixmatch$ nearly matches the fully-supervised performance on the same training set almost immediately -- for example, $\mixmatch$ achieves an error rate of $2.22\%$ with only 250 labels on SVHN+Extra compared to the fully-supervised performance of $1.71\%$.
Interestingly, on SVHN+Extra $\mixmatch$ outperformed fully supervised training on SVHN without \textit{extra} ($2.59\%$ error) for every labeled data amount considered.
To emphasize the importance of this, consider the following scenario:
You have $73257$ examples from SVHN with $250$ examples labeled and are given a choice: You can either obtain $8\times$ more unlabeled data and use $\mixmatch$ or obtain $293\times$ more labeled data and use fully-supervised learning.
Our results suggest that obtaining additional unlabeled data and using $\mixmatch$ is more effective, which conveniently is likely much cheaper than obtaining $293\times$ more labels.

\begin{table}
\centering
\begin{tabular}{lrrrrrr}
\toprule
Labels & $250$ & $500$ & $1000$ & $2000$ & $4000$ & All \\
\midrule
 SVHN & $3.78 \pm 0.26$ & $3.64 \pm 0.46$ & $3.27 \pm 0.31$ & $3.04 \pm 0.13$ & $2.89 \pm 0.06$ & $2.59$ \\
 SVHN+Extra & $2.22 \pm 0.08$ & $2.17 \pm 0.07$ & $2.18 \pm 0.06$ & $2.12 \pm 0.03$ & $2.07 \pm 0.05$ & $1.71$ \\
\bottomrule
\end{tabular}
\vskip 0.1in
\caption{Comparison of error rates for SVHN and SVHN+Extra for $\mixmatch$. The last column (``All'') contains the fully-supervised performance with all labels in the corresponding training set.}
\label{tab:svhn_nox}
\vskip -0.2in
\end{table}

\paragraph{STL-10}
STL-10 contains $5000$ training examples aimed at being used with $10$ predefined folds (we use the first 5 only) with $1000$ examples each.
However, some prior work trains on all $5000$ examples.
We thus compare in both experimental settings.
With $1000$ examples $\mixmatch$ surpasses both the state-of-the-art for $1000$ examples as well as the state-of-the-art using all $5000$ labeled examples.
Note that none of the baselines in \cref{tab:stl10} use the same experimental setup (i.e. model), so it is difficult to directly compare the results; however, because $\mixmatch$ obtains the lowest error by a factor of two, we take this to be a vote in confidence of our method.
We used $\lambda_{\mathcal{U}}=50$.

\subsubsection{Ablation Study}
\label{sec:ablation}

Since $\mixmatch$ combines various semi-supervised learning mechanisms, it has a good deal in common with existing methods in the literature.
As a result, we study the effect of removing or adding components in order to provide additional insight into what makes $\mixmatch$ performant.
Specifically, we measure the effect of
\begin{itemize}
    \item using the mean class distribution over $K$ augmentations or using the class distribution for a single augmentation (i.e.\ setting $K = 1$)
    \item removing temperature sharpening (i.e.\ setting $T = 1$)
    \item using an exponential moving average (EMA) of model parameters when producing guessed labels, as is done by Mean Teacher \cite{tarvainen2017weight}
    \item performing $\mixup$ between labeled examples only, unlabeled examples only, and without mixing across labeled and unlabeled examples
    \item using Interpolation Consistency Training \cite{verma2019interpolation}, which can be seen as a special case of this ablation study where only unlabeled mixup is used, no sharpening is applied and EMA parameters are used for label guessing.
\end{itemize}
We carried out the ablation on CIFAR-10 with $250$ and $4000$ labels; the results are shown in \cref{tab:ablation}.
We find that each component contributes to $\mixmatch$'s performance, with the most dramatic differences in the $250$-label setting.
Despite Mean Teacher's effectiveness on SVHN (\cref{fig:vary_svhn}), we found that using a similar EMA of parameter values hurt $\mixmatch$'s performance slightly.

\begin{table}
\centering
\begin{tabular}{lrr}
\toprule
Ablation & $250$ labels & $4000$ labels \\
\midrule
$\mixmatch$ & $11.80$ & $6.00$ \\
$\mixmatch$ without distribution averaging ($K = 1$) & $17.09$ & $8.06$  \\
$\mixmatch$ with $K = 3$ & $11.55$ & $6.23$  \\
$\mixmatch$ with $K = 4$ & $12.45$ & $5.88$  \\
$\mixmatch$ without temperature sharpening ($T = 1$) & $27.83$ & $10.59$ \\
$\mixmatch$ with parameter EMA & $11.86$ & $6.47$ \\
$\mixmatch$ without $\mixup$ & $39.11$ & $10.97$ \\
$\mixmatch$ with $\mixup$ on labeled only & $32.16$ & $9.22$ \\
$\mixmatch$ with $\mixup$ on unlabeled only & $12.35$ & $6.83$ \\
$\mixmatch$ with $\mixup$ on separate labeled and unlabeled & $12.26$ & $6.50$ \\
Interpolation Consistency Training \cite{verma2019interpolation} & $38.60$ & $6.81$ \\
\bottomrule
\end{tabular}
\vskip 0.1in
\caption{Ablation study results. All values are error rates on CIFAR-10 with $250$ or $4000$ labels.
}
\label{tab:ablation}
\vskip -0.2in
\end{table}

\subsection{Privacy-Preserving Learning and Generalization}
\label{sec:dp_experiments}

Learning with privacy allows us to measure our approach's ability to generalize. Indeed, protecting the privacy of training data amounts to proving that the model does not overfit: a learning algorithm is said to be differentially private (the most widely accepted technical definition of privacy) if adding, modifying, or removing any of its training samples is guaranteed not to result in a statistically significant difference in the model parameters learned~\cite{dwork2016calibrating}. For this reason, learning with differential privacy is, in practice, a form of regularization~\cite{nissim2015generalization}.
Each training data access constitutes a potential privacy leakage, encoded as the pair of the input and its label. Hence, approaches for deep learning from private training data, such as DP-SGD~\citep{abadi2016deep} and PATE~\citep{papernot2016semi}, benefit from accessing as few labeled private training points as possible when computing updates to the model parameters. Semi-supervised learning is a natural fit for this setting.

We use the PATE framework for learning with privacy. A student is trained in a semi-supervised way from public \textit{unlabeled} data, part of which is labeled by an ensemble of teachers with access to private \textit{labeled} training data. The fewer labels a student requires to reach a fixed accuracy, the stronger is the privacy guarantee it provides. Teachers use a noisy voting mechanism to respond to label queries from the student, and they may choose \textit{not} to provide a label when they cannot reach a sufficiently strong consensus.
For this reason, if $\mixmatch$ improves the performance of PATE, it would also illustrate $\mixmatch$'s improved generalization from few canonical exemplars of each class. 

We compare the accuracy-privacy trade-off achieved by $\mixmatch$ to a VAT~\cite{miyato2018virtual} baseline on SVHN. VAT achieved the previous state-of-the-art of $91.6\%$ test accuracy for a privacy loss of $\varepsilon=4.96$~\cite{papernot2018scalable}.
Because $\mixmatch$ performs well with few labeled points, it is able to achieve $95.21 \pm 0.17\%$ test accuracy for a much smaller privacy loss of $\varepsilon=0.97$.
Because $e^\varepsilon$ is used to measure the degree of privacy, the  improvement is approximately $e^4\approx55\times$, a significant improvement. A privacy loss $\varepsilon$ below 1 corresponds to a much stronger privacy guarantee.
Note that in the private training setting the student model only uses 10,000 total examples.

\section{Conclusion}

We introduced $\mixmatch$, a semi-supervised learning method which combines ideas and components from the current dominant paradigms for SSL.
Through extensive experiments on semi-supervised and privacy-preserving learning, we found that $\mixmatch$ exhibited significantly improved performance compared to other methods in all settings we studied, often by a factor of two or more reduction in error rate.
In future work, we are interested in incorporating additional ideas from the semi-supervised learning literature into hybrid methods and continuing to explore which components result in effective algorithms.
Separately, most modern work on semi-supervised learning algorithms is evaluated on image benchmarks; we are interested in exploring the effectiveness of $\mixmatch$ in other domains.

\ifsubmission
\else
\subsubsection*{Acknowledgement}
We would like to thank Balaji Lakshminarayanan for his helpful theoretical insights.
\fi

\bibliography{biblio}
\bibliographystyle{plain}

\appendix

\newpage
\section{Notation and definitions}
{\footnotesize 
\begin{tabular}{p{1in}p{4in}}
\toprule
\textbf{Notation} & \textbf{Definition} \\
\midrule
$\xent(p,q)$ & Cross-entropy between ``target'' distribution $p$ and ``predicted'' distribution $q$\\
\midrule
$x$ & A labeled example, used as input to a model \\
\midrule
$p$ & A (one-hot) label\\
\midrule
$L$ & The number of possible label classes (the dimensionality of $p$) \\
\midrule
$\mathcal{X}$ & A batch of labeled examples and their labels \\
\midrule
$\mathcal{X}^\prime$ & A batch of processed labeled examples produced by $\mixmatch$ \\
\midrule
$u$ & An unlabeled example, used as input to a model \\
\midrule
$q$ & A guessed label distribution for an unlabeled example \\
\midrule
$\mathcal{U}$ & A batch of unlabeled examples \\
\midrule
$\mathcal{U}^\prime$ & A batch of processed unlabeled examples with their label guesses produced by $\mixmatch$ \\
\midrule
$\theta$ & The model's parameters \\
\midrule
$\pmodel(y \mid x; \theta)$ & The model's predicted distribution over classes \\
\midrule
$\augment(x)$ &  A stochastic data augmentation function that returns a modified version of $x$. For example, $\augment(\cdot)$ could implement randomly shifting an input image, or implement adding a perturbation sampled from a Gaussian distribution to $x$.\\
\midrule
$\lambda_\mathcal{U}$ & A hyper-parameter weighting the contribution of the unlabeled examples to the training loss\\
\midrule
$\alpha$ & Hyperparameter for the $\betadist$ distribution used in $\mixup$ \\
\midrule
$T$ & Temperature parameter for sharpening used in $\mixmatch$ \\
\midrule
$K$ & Number of augmentations used when guessing labels in $\mixmatch$ \\
\midrule
\end{tabular}
}

\newpage
\section{Tabular results}
\subsection{CIFAR-10}
Training the same model with supervised learning on the entire $50000$-example training set achieved an error rate of $4.13\%$.
\begin{table}[H]
\centering
\begin{tabular}{lrrrrr}
\toprule
Methods/Labels & 250 & 500 & 1000 & 2000 & 4000 \\
\midrule
PiModel & $53.02 \pm 2.05$ & $41.82 \pm 1.52$ & $31.53 \pm 0.98$ & $23.07 \pm 0.66$ & $17.41 \pm 0.37$ \\
PseudoLabel & $49.98 \pm 1.17$ & $40.55 \pm 1.70$ & $30.91 \pm 1.73$ & $21.96 \pm 0.42$ & $16.21 \pm 0.11$ \\
Mixup & $47.43 \pm 0.92$ & $36.17 \pm 1.36$ & $25.72 \pm 0.66$ & $18.14 \pm 1.06$ & $13.15 \pm 0.20$ \\
VAT & $36.03 \pm 2.82$ & $26.11 \pm 1.52$ & $18.68 \pm 0.40$ & $14.40 \pm 0.15$ & $11.05 \pm 0.31$ \\
MeanTeacher & $47.32 \pm 4.71$ & $42.01 \pm 5.86$ & $17.32 \pm 4.00$ & $12.17 \pm 0.22$ & $10.36 \pm 0.25$ \\
MixMatch & $11.08 \pm 0.87$ & $9.65 \pm 0.94$ & $7.75 \pm 0.32$ & $7.03 \pm 0.15$ & $6.24 \pm 0.06$ \\
\bottomrule
\end{tabular}
\vskip 0.1in
\caption{Error rate (\%) for CIFAR10.}
\label{tab:cifar10}
\end{table}

\subsection{SVHN}
Training the same model with supervised learning on the entire $73257$-example training set achieved an error rate of $2.59\%$.
\begin{table}[H]
\centering
\begin{tabular}{lrrrrr}
\toprule
Methods/Labels & 250 & 500 & 1000 & 2000 & 4000 \\
\midrule
PiModel & $17.65 \pm 0.27$ & $11.44 \pm 0.39$ & $8.60 \pm 0.18$ & $6.94 \pm 0.27$ & $5.57 \pm 0.14$ \\
PseudoLabel & $21.16 \pm 0.88$ & $14.35 \pm 0.37$ & $10.19 \pm 0.41$ & $7.54 \pm 0.27$ & $5.71 \pm 0.07$ \\
Mixup & $39.97 \pm 1.89$ & $29.62 \pm 1.54$ & $16.79 \pm 0.63$ & $10.47 \pm 0.48$ & $7.96 \pm 0.14$ \\
VAT & $8.41 \pm 1.01$ & $7.44 \pm 0.79$ & $5.98 \pm 0.21$ & $4.85 \pm 0.23$ & $4.20 \pm 0.15$ \\
MeanTeacher & $6.45 \pm 2.43$ & $3.82 \pm 0.17$ & $3.75 \pm 0.10$ & $3.51 \pm 0.09$ & $3.39 \pm 0.11$ \\
MixMatch & $3.78 \pm 0.26$ & $3.64 \pm 0.46$ & $3.27 \pm 0.31$ & $3.04 \pm 0.13$ & $2.89 \pm 0.06$ \\
\bottomrule
\end{tabular}
\vskip 0.1in
\caption{Error rate (\%) for SVHN.}
\label{tab:svhn}
\end{table}

\newpage
\subsection{SVHN+Extra}
Training the same model with supervised learning on the entire $604388$-example training set achieved an error rate of $1.71\%$.
\begin{table}[H]
\centering
\begin{tabular}{lrrrrr}
\toprule
Methods/Labels & 250 & 500 & 1000 & 2000 & 4000 \\
\midrule
PiModel & $13.71 \pm 0.32$ & $10.78 \pm 0.59$ & $8.81 \pm 0.33$ & $7.07 \pm 0.19$ & $5.70 \pm 0.13$ \\
PseudoLabel & $17.71 \pm 0.78$ & $12.58 \pm 0.59$ & $9.28 \pm 0.38$ & $7.20 \pm 0.18$ & $5.56 \pm 0.27$ \\
Mixup & $33.03 \pm 1.29$ & $24.52 \pm 0.59$ & $14.05 \pm 0.79$ & $9.06 \pm 0.55$ & $7.27 \pm 0.12$ \\
VAT & $7.44 \pm 1.38$ & $7.37 \pm 0.82$ & $6.15 \pm 0.53$ & $4.99 \pm 0.30$ & $4.27 \pm 0.30$ \\
MeanTeacher & $2.77 \pm 0.10$ & $2.75 \pm 0.07$ & $2.69 \pm 0.08$ & $2.60 \pm 0.04$ & $2.54 \pm 0.03$ \\
MixMatch & $2.22 \pm 0.08$ & $2.17 \pm 0.07$ & $2.18 \pm 0.06$ & $2.12 \pm 0.03$ & $2.07 \pm 0.05$ \\
\bottomrule
\end{tabular}
\vskip 0.1in
\caption{Error rate (\%) for SVHN+Extra.}
\label{tab:svhn_extra}
\end{table}

\begin{figure}[H]
    \centering
    \includegraphics[width=\textwidth]{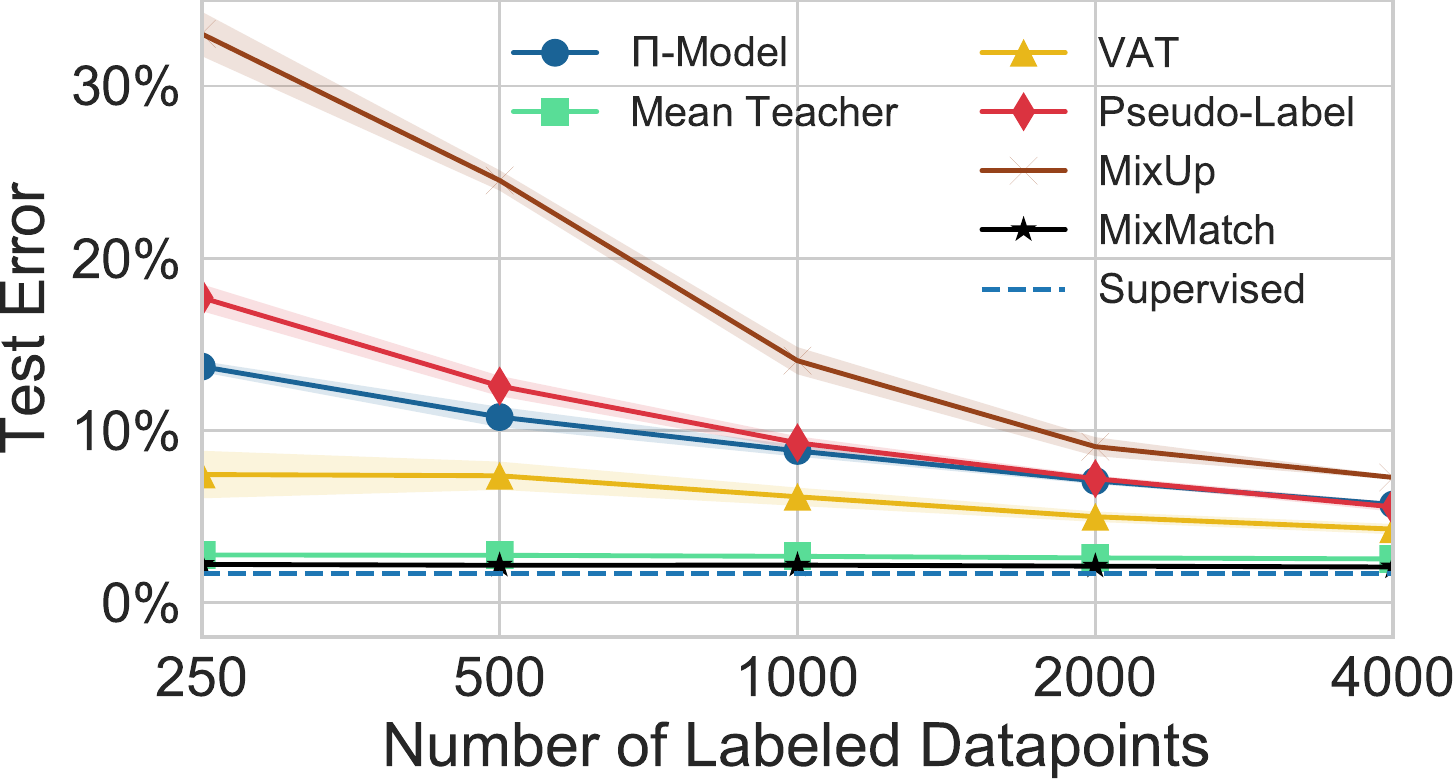}
    \caption{Error rate comparison of MixMatch to baseline methods on SVHN+Extra for
    a varying number of labels. With $250$ examples we reach nearly the state of the art compared to supervised training for this model.}
    \label{fig:vary_svhn_extra}
\end{figure}

\section{13-layer ConvNet results}

Early work on semi-supervised learning used a 13-layer convolutional network architecture \cite{miyato2018virtual,tarvainen2017weight,laine2016temporal}.
In \cref{tab:conv13} we present results on a similar architecture.
We caution against comparing these numbers directly to previous work as we use a different implementation and training process \cite{oliver2018realistic}.

\begin{table}[H]
\footnotesize
\begin{center}
\begin{tabular}{lrrrr}
\toprule
Method & \multicolumn{2}{c}{CIFAR-10} & \multicolumn{2}{c}{SVHN} \\
& 250 & 4000 & 250 & 1000 \\ 
\midrule
Mean Teacher & 46.34 & 88.57 &94.00 &96.00 \\
\textbf{MixMatch} & \textbf{85.69} & \textbf{93.16} & \textbf{96.41} & \textbf{96.61} \\
\bottomrule
\end{tabular}
\end{center}
\caption{Results on a $13$-layer convolutional network architecture.}\label{tab:conv13}
\end{table}

\end{document}